\documentclass[authoryear,preprint,review,12pt]{elsarticle}

\usepackage{graphics} 

\usepackage{float}
\usepackage{graphicx}
\usepackage{wrapfig} 

\usepackage[draft]{todonotes}   

\usepackage[labelfont=bf]{caption}
\usepackage{caption}
\usepackage{color}
\usepackage{amsmath} 
\usepackage{mathtools} 
\usepackage{amsfonts} 
\usepackage{amssymb}
\usepackage{array,longtable}

\usepackage{array}
\usepackage{booktabs}
\usepackage{rotating}
\usepackage{multirow}
\usepackage{multicol}
\usepackage{lscape}
\usepackage{subfig}

\usepackage[utf8]{inputenc}
\usepackage[T1]{fontenc}
\usepackage{textcomp}
\usepackage{gensymb}

\usepackage[export]{adjustbox}

\usepackage{algorithm,algorithmicx,algpseudocode}

\usepackage{color}

\usepackage[colorlinks=true,linkcolor=blue,citecolor=blue]{hyperref}

\usepackage{comment}

\usepackage[draft]{todonotes}   

\usepackage{soul}

\definecolor{darkgreen}{rgb}{0.53, 0.66, 0.42}
\definecolor{azure}{rgb}{0.0, 0.5, 1.0}

\journal{} 

\begin{document}

\begin{frontmatter}



\title{Quantifying the Reproducibility of Graph Neural Networks using Multigraph Brain Data}

\author{Mohammed Amine Gharsallaoui \fnref{BASIRA}}
\author{Islem Rekik \corref{cor} \fnref{BASIRA,DUNDEE}}
\author{and for the Alzheimer's Disease Neuroimaging Initiative}

\cortext[cor]{Corresponding author: irekik@itu.edu.tr; \url{http://basira-lab.com/}. Data used in preparation of this article were obtained from the Alzheimer's Disease Neuroimaging Initiative (ADNI) database (\url{adni.loni.usc.edu}). As such, the investigators within the ADNI contributed to the design and implementation of ADNI and/or provided data but did not participate in analysis or writing of this report. A complete listing of ADNI investigators can be found at: http://adni.loni.usc.edu/wp- content/uploads/how to apply/ADNI Acknowledgement List.pdf}

\address[BASIRA]{BASIRA lab, Faculty of Computer and Informatics Engineering, Istanbul Technical University, Istanbul, Turkey}
\address[DUNDEE]{School of Science and Engineering, Computing, University of Dundee, UK \ }

\begin{abstract}
Graph neural networks (GNNs) have witnessed an unprecedented proliferation in tackling several problems in computer vision, computer-aided diagnosis and related fields. While prior studies have focused on boosting the model accuracy, quantifying the reproducibility of the most discriminative features identified by GNNs is still an intact problem that yields concerns about their reliability in clinical applications in particular. Specifically, the reproducibility of biological markers across clinical datasets and distribution shifts across classes (e.g., healthy and disordered brains) is of paramount importance in revealing the underpinning mechanisms of diseases as well as propelling the development of personalized treatment. Motivated by these issues, we propose, for the first time, reproducibility-based GNN selection (RG-Select), a framework for GNN reproducibility assessment via the quantification of the most discriminative features (i.e., biomarkers) shared between different models. To ascertain the soundness of our framework, the reproducibility assessment embraces variations of different factors such as training strategies and data perturbations. Despite these challenges, our framework successfully yielded replicable conclusions across different training strategies and various clinical datasets. Our findings could thus pave the way for the development of biomarker trustworthiness and reliability assessment methods for computer-aided diagnosis and prognosis tasks. RG-Select code is available on GitHub at \url{https://github.com/basiralab/RG-Select}.

\end{abstract}


\end{frontmatter}


\section{Introduction}


The scope of deep learning (DL) application in neuroscience is marking an exponential growth in many directions thanks to its proven efficiency in tackling many problems such as classification \citep{richards2019deep} or regression \citep{smith2013comparison}. The abundance of non-invasive neuroimaging datasets acquired from different modalities (e.g., structural and functional MRI) and the availability of new computational frameworks are indubitably pushing the boundaries of research towards deepening our understanding of brain connectivity \citep{bassett2017network}. In network neuroscience, in particular, the graph structure is considered as a powerful data representation thanks to its capacity in encoding connections between different brain regions \citep{van2019cross, deletoile2017graph, farahani2019application, he2010graph}. In fact, a brain connectome is a map of connections in the brain wiring different anatomical regions of interest (ROIs), providing a comprehensive map of the network structure of the brain. Thus, it helps better understand the anatomically based interactions between different ROIs \citep{toga2012mapping} in a non-invasive manner. A brain connectome can be modeled as a graph where each node  denotes an ROIs and an edge connects two ROIs quantifying their interaction. Applying traditional DL frameworks to graphs does not lead to satisfactory results due to their incapacity in exploiting the topological properties of such non-Euclidian data \citep{bronstein2017geometric,henaff2015deep}. To mitigate this limitation, graph neural networks (GNNs), an extended family of DL methods dealing with non-Euclidian data, have been proposed as an alternative to traditional DL algorithms in many fields \citep{zhang2020deep, wang2019deep, monti2017geometric} including the field of network neuroscience \citep{Bessadok:2021,wang2021generalizable}. GNNs have demonstrated a promising potential in capturing the topological features of graphs to perform a given task such as classification or regression \citep{zhou2020graph, wu2020comprehensive, xu2018powerful}. 

Thus far, most DL and GNN classification models applied in network neuroscience have focused on increasing the accuracy in discriminating between two neurological states (e.g., healthy and neurologically disordered) \citep{Bessadok:2021,rashid2016classification,shirer2012decoding,alper2013weighted}. Notably, instead of evaluating the efficiency in discriminating between two classes, GNNs can be evaluated in their capacity to reproduce the most reliable set of discriminative ROIs in a given learning task. Specifically, given a set of models, one can quantify how likely can a model identify the most discriminative ROIs shared across models. Hence, considering an input dataset with two classes (e.g., healthy \emph{vs} disordered), two models are \emph{reproducible} if they identify the same top discriminative biomarkers in the original input space across different data perturbation strategies. The comparison of learning models based on accuracy does not necessarily compare the sets of identified biomarkers across models. In addition to being agnostic to the reliability of biomarkers, such accuracy-based comparison only focuses on the final classification and does not take into account the learned weights of each model. Although this might produce good classification results, it can be criticized as being uninformative about biomarker reliability and reproducibility. Besides, this might hinder clinical understandability since the identification of the biomarkers underlying a specific brain disorder is crucial for clinical interpretability \citep{povero2020characterization}.

Although still under discussion, reproducibility has been suggested to reflect and investigate the biomarkers' reliability of classification models \citep{georges2020identifying}. In more detail, the predictions made by GNNs are learned by identifying the different brain connectivity alterations between brain regions that mark a particular disorder. To deepen our understanding of brain connectivity, quantifying the reproducibility of GNNs in terms of biomarkers becomes crucial to investigate more rigorously their reliability. In this context, the reproducibility of a model can be looked at as how likely it is \emph{congruent with other models}. Specifically, here we define the reproducibility score of a given GNN model based on the intersection of its most relevant features with feature sets identified by other GNN models. It is also important to note that since reliability is often correlated with reproducibility, the GNN assessment has to be generalized across various perturbations of the training and testing data distributions. 

\begin{figure}[h]
\centering
\includegraphics[width=\linewidth]{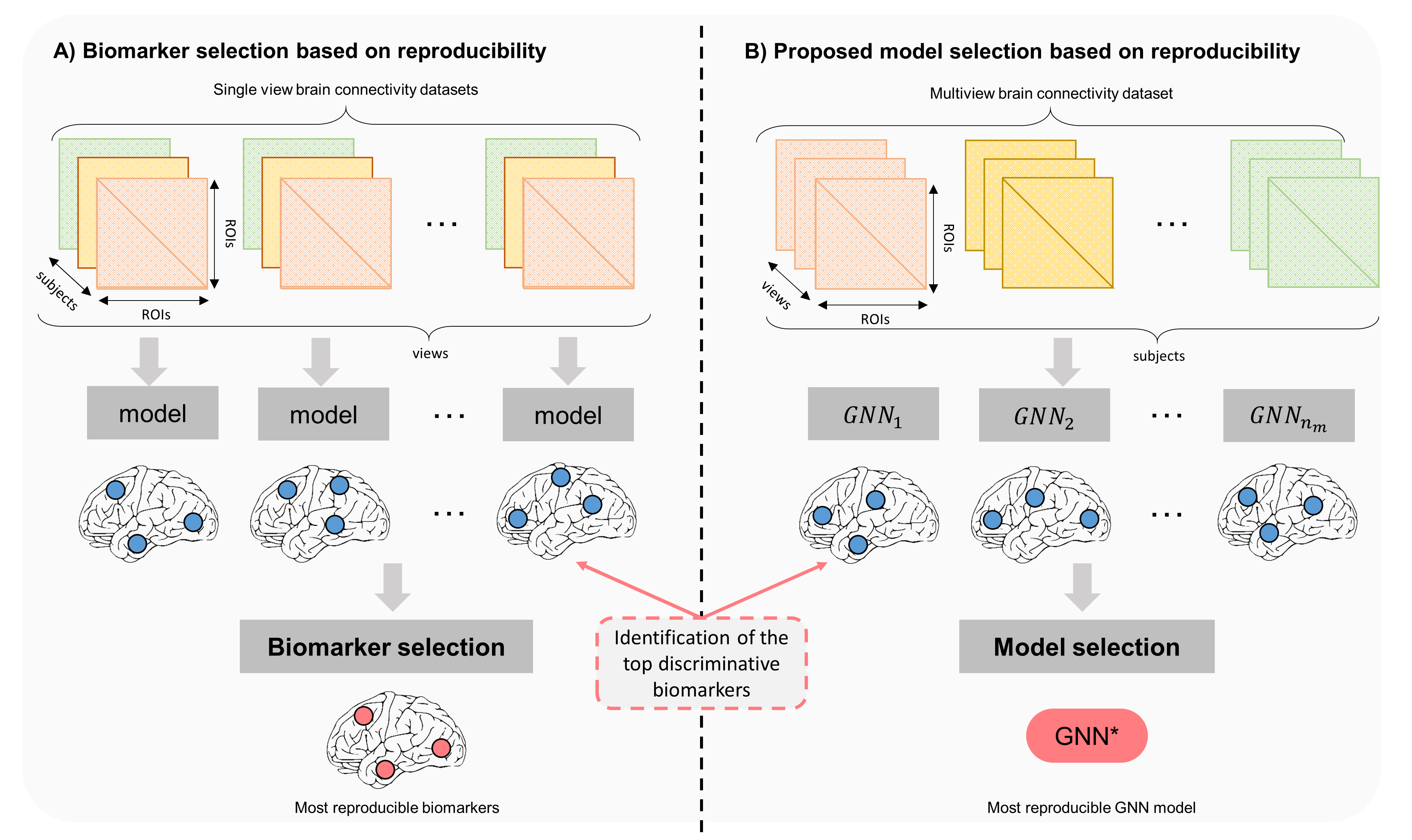}
\caption{Conventional accuracy-based model selection versus reproducibility-based model selection. \textbf{A)} Conventional approach focuses on selecting the model with the highest prediction accuracy. \textbf{B)} Our proposed approach aims to select the most reproducible model to produce reliable clinical diagnosis and prognosis.}
\label{fig:concept}
\end{figure}

A few studies have been proposed to tackle the problem of biomarker or feature reproducibility. \citep{jin2020generalizable} worked on reproducibility across datasets collected from different sites to evaluate the generalizability and robustness of a given model. \citep{du2020neuromark} investigated the reproducibility of biomarkers across datasets to extract the most reproducible brain alterations responsible for a neurological abnormality. Although they have generated robust conclusions, such methods do not study the reproducibility across brain connectivity multigraph datasets (i.e., graphs with different connectivity measures for the same pair of nodes). Another line of works has focused on reproducibility across models \citep{georges2020identifying}. This approach reflects more robustness since it considers multiple models at once and takes into consideration datasets containing brain multigraphs \citep{dhifallah2020estimation,lisowska2017pairing,mahjoub2018brain}. However, the proposed framework in \citep{georges2020identifying} focused only on traditional feature selection (FS) methods and cannot be applied directly to GNNs due to their complexity. In fact, extracting the top biomarkers in FS methods is inherently straightforward, unlike other frameworks. Most GNNs include graph embeddings or graph reshaping operations which alter the original dimensions in the input space \citep{Bessadok:2021}. To overcome these limitations, in analogy with reproducibility of FS methods where the most discriminative features from the input space are selected, we can look at the weights learned in a deep learning model as an indicator of the discriminativeness for biomarkers (i.e., sample features in general). For this purpose, we extract the weights of a given GNN model in its last layer preserving the original graph dimensions. Consequently, we build a feature map for each GNN characterizing the discriminativeness assigned to the neurological biomarkers, which denote brain ROIs in our case. This choice is also justified by the fact that the last layer can be looked at as a weighted combination of all the previous neurons in a given neural network. Consequently, we analyze the intersection of the different GNN specific feature maps using different strategies with the aim of selecting the \emph{most reproducible GNN} as illustrated in \textbf{Fig.} \ref{fig:concept}.        

In this study, we propose the concept of reproducibility as a criterion for best GNN model selection. Notably, the feature map of weights respective to biomarkers reflects the importance accorded by a given GNN projected in the input domain. By conceptualizing the reproducibility in feature selection case as the consensus in terms of selected biomarkers across different models, we can extend this approach to the GNN models by regarding the learned weights as an importance factor. We can then pick the top-weighted biomarkers to investigate the overlap across GNNs from different angles. To ensure generalizability, our study implicates variations in multiple factors such as brain connectivity measures, training data distribution perturbation strategies and the number of top biomarkers to be considered for a given dataset of two different neurological states (e.g., healthy \emph{vs.} disordered). Depending on these factors, we target GNN reproducibility by using different techniques with the aim of establishing a generalizable and trustworthy clinical interpretation.

In view of these aims, we propose reproducibility based GNN selection (RG-Select)\footnote{\url{https://github.com/basiralab/RG-Select}}, a novel framework that investigates the reproducibility of GNN classifiers in datasets of brain connectivity multigraphs, where two nodes are connected by multiple edges, each capturing a particular facet of the brain interactivity. Specifically, we aim to rigorously assess our framework with different settings in order to provide generalizable results. In this context, our study incorporates the variations of the following factors: (1) GNNs, (2) brain connectivity measures per dataset, (3) training strategies, (4) number of top biomarkers to be selected, and (5) connectivity measures (e.g., cortical thickness and sulcal depth). Taking into account those factors and given a pool of GNN models and a particular dataset of interest, our RG-Select identifies the most reproducible GNN model.

\section{Proposed reproducibility based graph neural network selection (RG-Select)}
\begin{figure}[H]
\centering
\includegraphics[width=14cm]{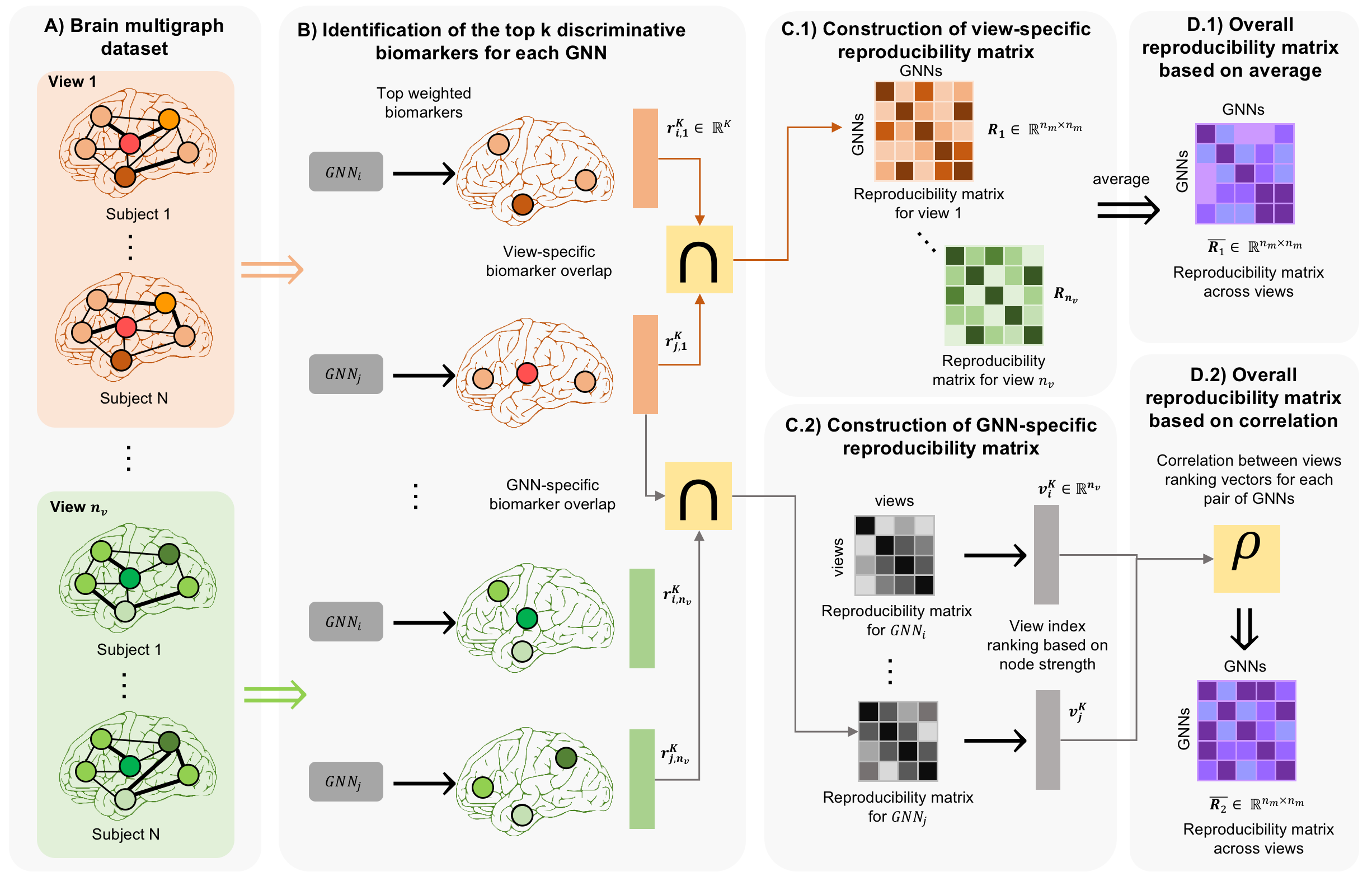}
\caption{Illustration of the proposed framework for GNN reproducibility assessment. A) We start with datasets of single-view graphs. B) We train different GNN models on the datasets. For each model-dataset combination, we extract the biomarkers absolute value weights vector. Then, we rank the resulting vectors to identify the top discriminative brain regions. C.1) We calculate the overlap ratio between resulting vectors from different GNNs and the same view. For each view, we build a reproducibility matrix. C.2) We compute the overlap ratio between resulting vectors from different views and the same GNN. Consequently, we obtain a matrix for each GNN. Next, we extract ranking vectors based on the node strength vectors resulting from these matrices. This step results in a ranking vector for each GNN. D.1) We compute the average of the resulting matrices to obtain the overall reproducibility matrix. D.2) We calculate the correlation between pairs of GNN resulting ranking vectors.}
\label{fig:main}
\end{figure}

In this section, we present in detail the proposed framework RG-Select for quantifying the reproducibility of GNNs as illustrated in \textbf{Fig.}~\ref{fig:main}. First, we construct single view datasets by separating the views in each multigraph. We train a set of GNNs on each dataset, separately. Following training, we extract a set of top discriminative biomarkers (i.e., ROIs) based on the ranking of their respective learned weights. In detail, we extract sets of top biomarkers with different sizes for generalizability purpose. Finally, we assign scores to each pair of models based on the inter-model discriminative biomarker overlap. Inter-model reproducibility scores will be used eventually to build the overall reproducibility matrix which incorporates variations of the different factors.
 
\subsection{Problem statement}
We denote $\mathcal{D}=(\mathcal{G},\mathcal{Y})$ as the dataset containing brain connectivity multigraphs with a set of classes respective to different brain neurological states to classify. Let $\mathcal{G} = \{\mathbf{G}_1, \mathbf{G}_2, \dots ,\mathbf{G}_{n}\}$ and $\mathcal{Y}=\{y_1, y_2, \dots ,y_n\}$ denote the set of the brain connectivity multigraphs and their labels, respectively. Each connectivity multigraph $\mathbf{G}_i$ is represented by a connectivity (adjacency) tensor $\mathbf{X}_i \in \mathbb{R}^{n_r \times n_r \times _v}$ and a label $y_i \in \{0, 1\}$ where $n_r$ and $n_v$ denote the number of ROIs and the number of views (i.e., edge types) of the brain multigraph, respectively. Let $\mathcal{D}^j=(\mathcal{G}^j,\mathcal{Y})$ be the dataset constructed from the $j^{th}$ view or measurement (e.g., cortical thickness) where $j \in \{1, ..., n_v\}$. We also denote a single view connectivity matrix as $\mathbf{X}_i^j \in \mathbb{R}^{n_r \times n_r}$, where $j \in \{1, ..., n_v\}$ is the view index in the multigraph. 

Given a pool of $n_m$ GNNs $\{{GNN}_1, {GNN}_2, \dots {GNN}_{n_m}\}$, we are interested in training a GNN model ${GNN}_i : \mathcal{G} \rightarrow \mathcal{Y}$ on the separate single view dataset $\{\mathcal{D}^j\}_{j=1}^{n_v}$. We aim to identify the best GNN that reproduces the same biomarkers differentiating between two brain states against different data perturbation strategies. Thus, we extract the weight vector $\mathbf{w}_i \in \mathbb{R}^{n_r}$ learned by the $i^{th}$ GNN model, where $i \in \{1, 2, \dots ,n_m \}$ in each experiment.  For a given dataset, we extract the weights for all the views and GNNs. Next, we rank the biomarkers based on the absolute value of their respective weights. Finally, we compute the reproducibility scores as detailed in what follows. 

\subsection{Model selection and evaluation}

Consistent with previous machine learning practices, we conducted separate model selection and evaluation steps to ensure fairness in the assessment of the models following the protocol detailed in \citep{errica2019fair}. To do so, we partition our training set into an inner training set and holdout subset. Next, we train the GNN on the inner training set and validate it on the holdout subset for the model selection. The model selection aims to tune the hyperparameters based on the performance on the validation set. Next, we select the optimal hyperparameters combination that brought the best results on the validation set. We then use the optimal hyperparameters in the model evaluation step where each model is assessed on a separate test set depending on the different $k$-fold cross-validation (CV). $k$-fold CV consists of $k$ different training/test splits used to evaluate the performance of the model. In each iteration, the model is tested on a subset of samples never used in the model selection step where the model is trained. We also ensure label stratification in the different data partitions so that the class proportions are preserved across all the training/test/validation splits. This protocol is motivated by the fact that there are some issues about the separation between model \emph{selection} and \emph{assessment} in different state-of-the-art GNN official implementations leading to unfair and biased comparisons \citep{errica2019fair}. 

\subsection{GNN training modes} 
We used different modes of training for our GNNs to ensure the generalizability of the results. We conducted resourceful training which is based on the conventional $k$-fold cross-validation protocol. It trains the models on the training set following the same fairness diagram detailed in \citep{errica2019fair}. Moreover, we used frugal training which follows a few-shot learning approach. This mode performs the model training on only a few samples and the evaluation on all the remaining subjects in the dataset. Using both methods is important to ensure that the results of our framework are agnostic to data perturbations and training strategies.

\subsection{Biomarker selection}
 
In contrast with the conventional approach which focuses on accuracy evaluation of a given classifier, here, we focus on model reproducibility in top discriminative features (i.e., biomarkers). Typically, the extraction of the most discriminative biomarkers for FS methods is straightforward. However, GNN methods have different architectures which makes it hard to implement a generalized way to extract the most important biomarkers. To circumvent this issue, we extract the weights of the last layer preserving the dimensionality of the input data (i.e., having the same number of features/ROIs). Specifically, given $n_r$ ROIs for each brain connectome, we rank these biomarkers based on their learned weights by the selected GNN. Based on that ranking, we extract $\mathbf{r}^{K_h}_{i,j} \in \mathbb{R}^{K_h}$ the vector containing the top $K_h$ biomarkers based on the weights learned by the $i^{th}$ GNN trained on the $j^{th}$ view of the input multigraph dataset.  

\textbf{\emph{Definition 1.}} \emph{Let $\mathbf{r}^{k}_{i,v}, \mathbf{r}^{k}_{j,v} \in \mathbb{R}^{n_r}$ denote two the vectors containing the top $k$ biomarkers learned on the same view $v$ by $GNN_i$ and $GNN_j$, respectively. We denote $r^{k}_{i,v}$ and $r^{k}_{j,v}$ as the two sets containing the regions included in $\mathbf{r}^{k}_{i,v}, \mathbf{r}^{k}_{j,v}$, respectively. We define the view-specific reproducibility on the view $v$ at threshold $k$ between models $i$ and $j$ as: \(p_v\) $(\mathbf{r}^{k}_{i,v}, \mathbf{r}^{k}_{j,v}) = \frac{|r^{k}_{i,v} \cap r^{k}_{j,v}|}{k}$}

\textbf{\emph{Definition 2.}} \emph{Let $\mathbf{r}^{k}_{g,i}, \mathbf{r}^{k}_{g,j} \in \mathbb{R}^{n_r}$ denote two the vectors containing the top $k$ biomarkers learned by the same ${GNN}_g$ on the views $i$ and $j$, respectively. We denote $r^{k}_{g,i}$ and $r^{k}_{g,j}$ as the two sets containing the regions included in $\mathbf{r}^{k}_{g,i}, \mathbf{r}^{k}_{g,j}$, respectively. We define the GNN-specific reproducibility by ${GNN}_g$ at threshold $k$ between views $i$ and $j$ as: \(p_g\) $(\mathbf{r}^{k}_{g,i}, \mathbf{r}^{k}_{g,j}) = \frac{|r^{k}_{g,i} \cap r^{k}_{g,j}|}{k}$}

\subsection{View-specific reproducibility matrix} 
For a pool containing $n_m$ GNNs, we aim to quantify the reproducibility between each pair of models. Since reproducibility reflects the commonalities between two sets of biomarkers, we propose to compute the ratio of the overlapping ROIs. First, we need to quantify the reproducibilities in the same domain (i.e., the same view). In other terms, for a given view $v$ and a threshold $K_h$ we calculate the ratio \(p_v\)$(r^{K_h}_{i,v}, r^{K_h}_{j,v})$ for each pair of models ${GNN}_i$ and ${GNN}_j$. Having the reproducibility calculated for each pair of GNNs, we construct the reproducibility matrix $\mathbf{R}_{v}^{K_h} \in \mathbb{R}^{n_m \times n_m}$ where $\mathbf{R}_{v}^{K_h} (i,j) =  p_v(r^{K_h}_{i,v} r^{K_h}_{j,v})$. Next, we generate the average reproducibility matrix by merging all the reproducibility matrices across the different $p$ thresholds $\mathbf{R}_{v} (i,j) = \frac{\sum_{h=1}^{n_k} \mathbf{R}_{v}^{K_h}(i,j)}{n_k}$, where $n_k$ is the number of threshold values. Finally, after calculating the reproducibility locally (i.e., for each view) we need to get a general overview of the reproducibility across all views. Therefore, we average the resulting matrix over all the views and training modes (i.e., perturbation strategy).

\subsection{GNN-specific reproducibility matrix}
Another way to quantify reproducibility is to start from quantifying the commonalities across views for the same GNN. This is motivated by the fact that GNNs might have varying behaviors (i.e., different learned weight distributions) across different data views. For the same model, we measure the GNN-specific reproducibility between the different views of the dataset. For a given ${GNN}_{g}$, we construct the matrix $\mathbf{R}_{g}^{K_h} \in \mathbb{R}^{n_v \times n_v}$ where $\mathbf{R}_{g}^{K_h} (i,j) = p_g(r^{K_h}_{g,i}, r^{K_h}_{g,j})$. Then, we average over the thresholds, $\mathbf{R}_{g} (i,j) = \frac{\sum_{h=1}^{n_k} \mathbf{R}_{g}^{K_h} (i,j)}{n_k}$. Finally, we calculate the average of the GNN-specific reproducibility matrix for each model across all the different training modes. 

\section{Results}

\subsection{Evaluation datasets} 

We evaluated our reproducibility framework on a small-scale and a large-scale brain connectivity datasets. The first dataset (AD/LMCI) contains 77 subjects (41 subjects are diagnosed with Alzheimer's diseases (AD) and 36 diagnosed with Late Mild Cognitive Impairment (LMCI)) from the Alzheimer's Disease Neuroimaging Initiative (ADNI) database GO public dataset) \citep{weiner2010alzheimer}. The second dataset (ASD/NC) includes 300 subjects equally partitioned between autism spectral disorder (ASD) and normal control (NC) states extracted from Autism Brain Imaging Data Exchange ABIDE I public dataset \citep{di2014autism}. The connectivities were obtained using FreeSurfer \citep{fischl2012freesurfer} by constructing cortical morphological networks for each subject from structural T1-w MRI \citep{mahjoub2018brain,dhifallah2020estimation}. Next, both left and right cortical hemispheres (LH and RH) are parcellated into 35 cortical regions of interest (ROIs) using Desikan-Killiany cortical atlas, respectively \citep{desikan2006automated,lisowska2017pairing}. Both AD/LMCI (RH and LH) brain multigraphs are constructed using 4 cortical measures: maximum principal curvature, cortical thickness, sulcal depth and average curvature. As for the ASD/NC dataset, brain multigraphs are generated from six cortical attributes which are the same attributes used for AD/LMCI datasets in addition to cortical surface area and minimum principle area. Specifically, for each node ${ROI}_i$ and for each cortical attribute, we calculate the average cortical measurement $\overline{a}_i$ across all its vertices. The weight of the connectivity linking ${ROI}_i$ and ${ROI}_j$ is the absolute distance between their average cortical attributes: $|\overline{a}_i-\overline{a}_j|$.


\subsection{GNN models}
For our reproducibility framework, we used 5 state-of-the-art GNN architectures: DiffPool \citep{ying2018hierarchical}, GAT \citep{velivckovic2017graph}, GCN \citep{kipf2016semi}, SAGPool \citep{lee2019self} and g-U-Nets \citep{gao2019graph}. DiffPool performs a differential pooling to generate a learned hierarchical representation of an input graph. At each pooling layer, DiffPool learns how to make soft assignment of nodes into clusters which will be the nodes of the following layer \citep{ying2018hierarchical}. GAT and GCN are originally designed for node classification. Here, we adapt them to perform graph classification task. Therefore, we added a linear layer that projects the node scores into a global score for the whole graph. GAT learns different weights to the neighbourhood to perform the aggregation in the following layer. GCN sequentially learns convolution weights that encode neighbourhood features and local graph structure \citep{kipf2016semi}. SAGPool performs graph convolutions to learn pooling and unpooling of graphs based on self-attention \citep{lee2019self}. g-U-Nets is a U-shape based GNN combining multiple encoders and decoders that perform pooling and unpooling of the graph, respectively \citep{gao2019graph}.       

\subsection{Training settings and hyperparameters} We have used two different types of training in our experiments: resourceful and frugal. For the resourceful training, we trained our models in the conventional train/test approach. To do so, we have made 3-fold and 5-fold cross-validation strategies. In addition to the resourceful training approach based on the $k$- fold cross-validation, we also evaluated our experiments with a frugal training approach based on few-shot learning. Here, we only trained the model on 2 samples per class for each dataset. To ensure generalizability of the findings of the experiments, we made 100 runs with different randomizations so that the samples selected for the training will not be redundant. We also used 4 thresholds for the top biomarkers extraction which are 5, 10, 15 and 20. All the hyperparameters were selected using grid search. For all models, the learning rates ranged between 0.0001 and 0.001. For DiffPool, the hidden dimension, the output dimension, the assignment ratio and the number of convolution layers were equal to 256, 512, 0.1 and 3, respectively. For GAT, the numbers of hidden units and head attentions were equal to 8. For GCN, the number of hidden units is equal to 64. For g-U-Nets, the number of layers, hidden and convolution layer dimensions were equal to 3,  512 and 48, respectively. For SAGPool, the hidden dimension and the pooling ratio were equal to 256 and 0.5, respectively. 

\subsection{Overall reproducibility matrices}
\subsubsection{View-specific matrix based reproducibility}
To quantify the reproducibility across GNN models we used 4 different methods. The first method consists of calculating the average between the view-specific reproducibility matrices over all the views of the selected dataset. This method is intuitive in order to combine the information calculated for each view.

\subsubsection{GNN-specific matrix based reproducibility}
We rank the views based on the GNN-specific reproducibility matrix. For each GNN, we extract a vector indicating the ranks of the views. Next, we calculate the correlation coefficient across pairs of GNNs based on their respective reproducibility matrices. Consequently, we construct a reproducibility matrix containing pairwise relations between GNNs.   This method is useful to reflect how the GNN are likely to have the same behavior across views. 

\begin{figure}[H]
\centering
\includegraphics[width=\linewidth]{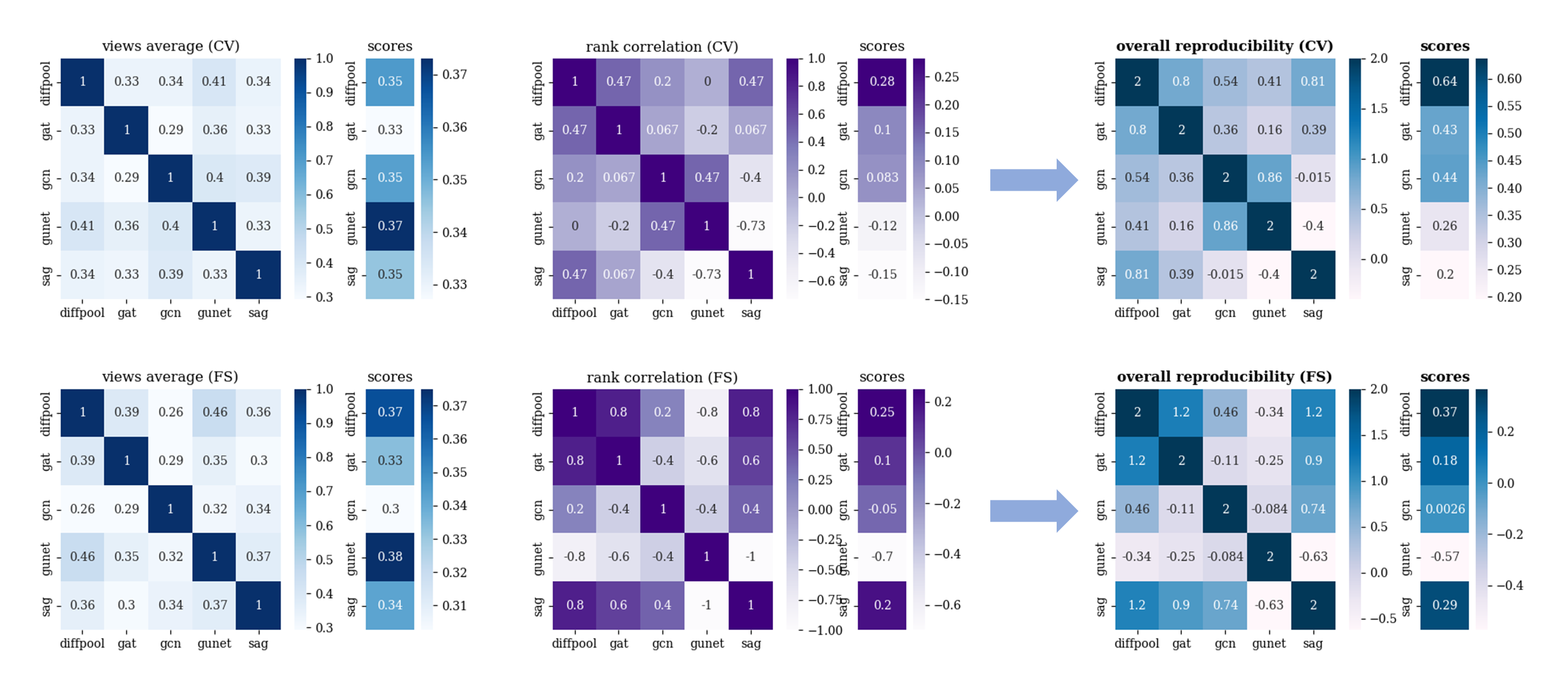}
\caption{\emph{Heatmaps of reproducibility matrices of AD/LMCI LH dataset.} The matrices were computed on cross-validation and few-shot training strategies, separately. AD: Alzheimer's disease. LMCI: late mild cognitive impairment. LH: left hemisphere. CV: cross-validation. FS: few-shot.}
\label{fig:hmaps_adlmcilh}
\end{figure}

\begin{figure}[H]
\centering
\includegraphics[width=\linewidth]{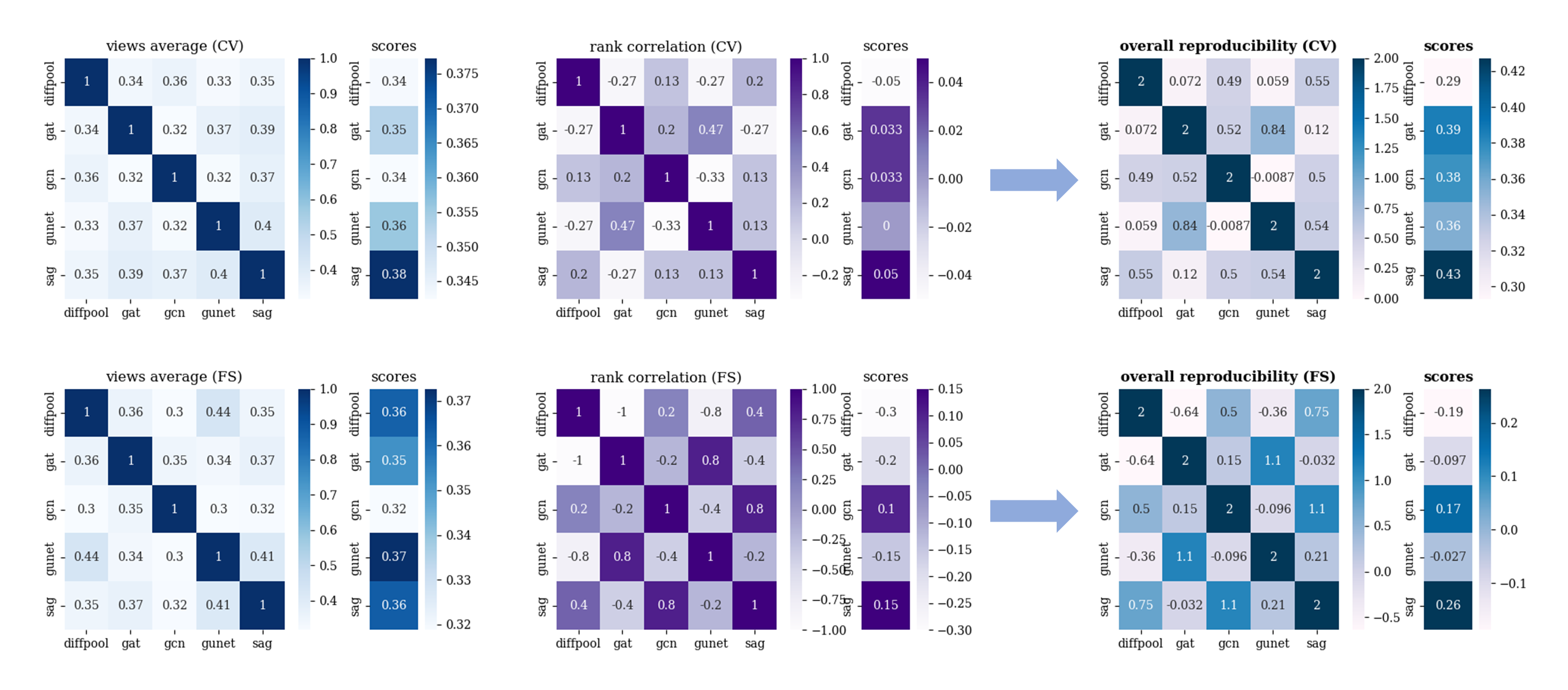}
\caption{\emph{Heatmaps of reproducibility matrices of AD/LMCI RH dataset}. The matrices were computed on cross-validation and few-shot training strategies, separately. AD: Alzheimer's disease. LMCI: late mild cognitive impairment. RH: right hemisphere. CV: cross-validation. FS: few-shot.}
\label{fig:hmaps_adlmcirh}
\end{figure}

\begin{figure}[H]
\centering
\includegraphics[width=\linewidth]{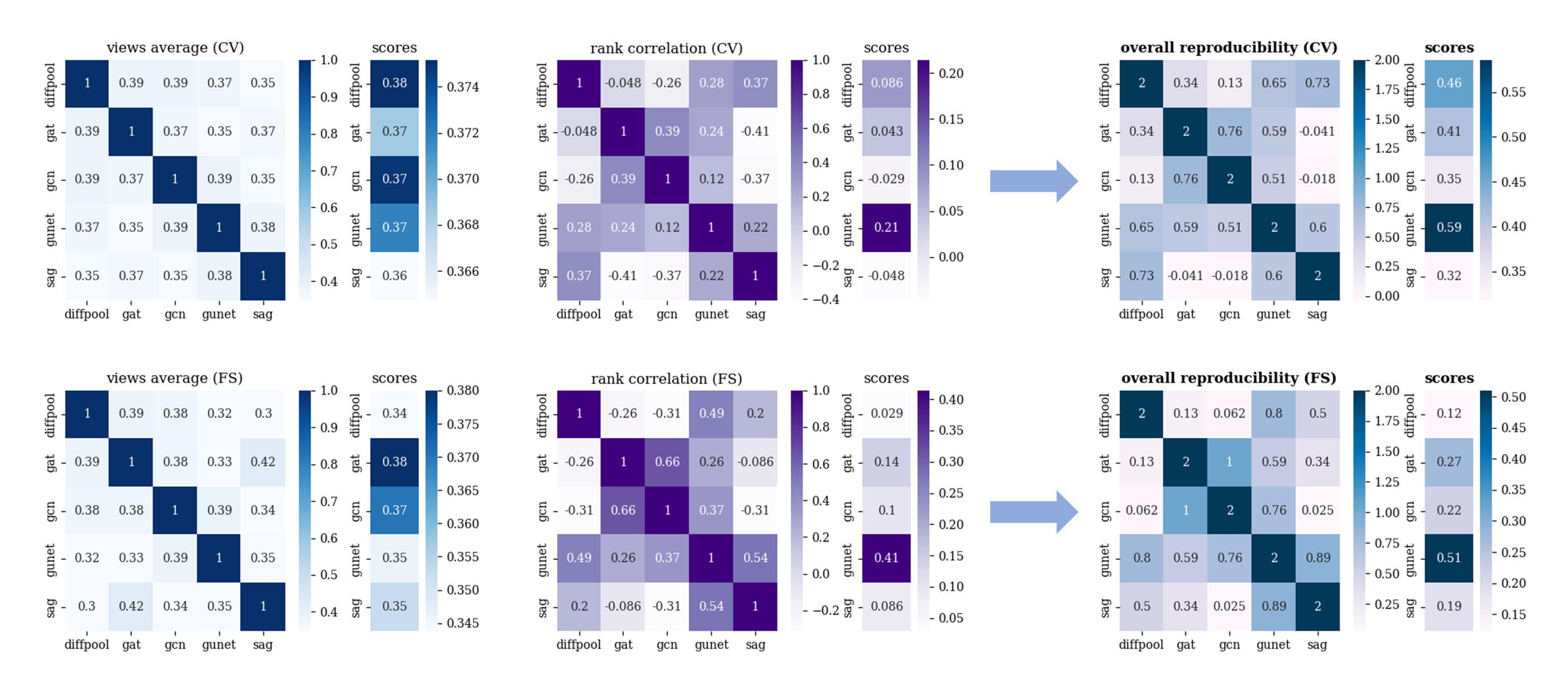}
\caption{\emph{Heatmaps of reproducibility matrices of ASD/NC LH dataset.} The matrices were computed on cross-validation and few-shot training strategies, separately. ASD: autism spectrum disorder. NC: normal control. LH: left hemisphere. CV: cross-validation. FS: few-shot.}
\label{fig:hmaps_asdnclh}
\end{figure}

\begin{figure}[H]
\centering
\includegraphics[width=\linewidth]{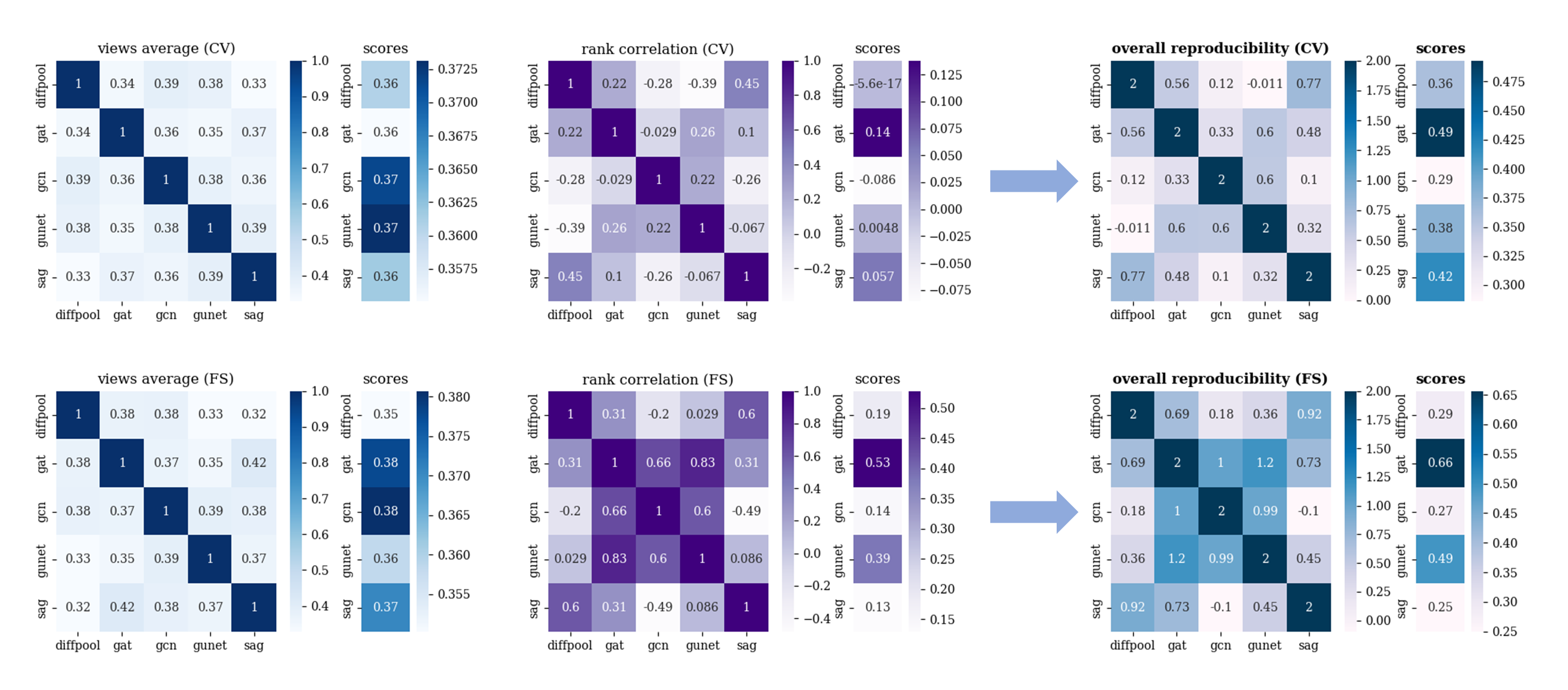}
\caption{\emph{Heatmaps of reproducibility matrices of ASD/NC RH dataset.} The matrices were computed on cross-validation and few-shot training strategies, separately. ASD: autism spectrum disorder. NC: normal control. RH: right hemisphere. CV: cross-validation. FS: few-shot.}
\label{fig:hmaps_asdncrh}
\end{figure}

\subsection{Most reproducible GNN selection}

In what follows, we define the reproducibility matrix as the summation of the two matrices detailed above. To take advantage of both GNN-specific and view-specific matrices, we sum the two reproducibility matrices detailed above. As such, we can look at the overall reproducibility matrix as a graph where the nodes represent the GNN models. Consequently, we use the node strength to quantify the reproducibility scores of the GNN models. This is conceptualized based on the intuition that the model reproducibility reflects the consensus in biomarkers with other models. Projecting this idea on the graph topology, node strength is a topological measure that encodes the magnitude of the connections with remaining entities in the graph. We define the most reproducible model as the node having the highest strength in the reproducibility graph. 

\begin{figure}[H]
\centering
\includegraphics[width=\linewidth]{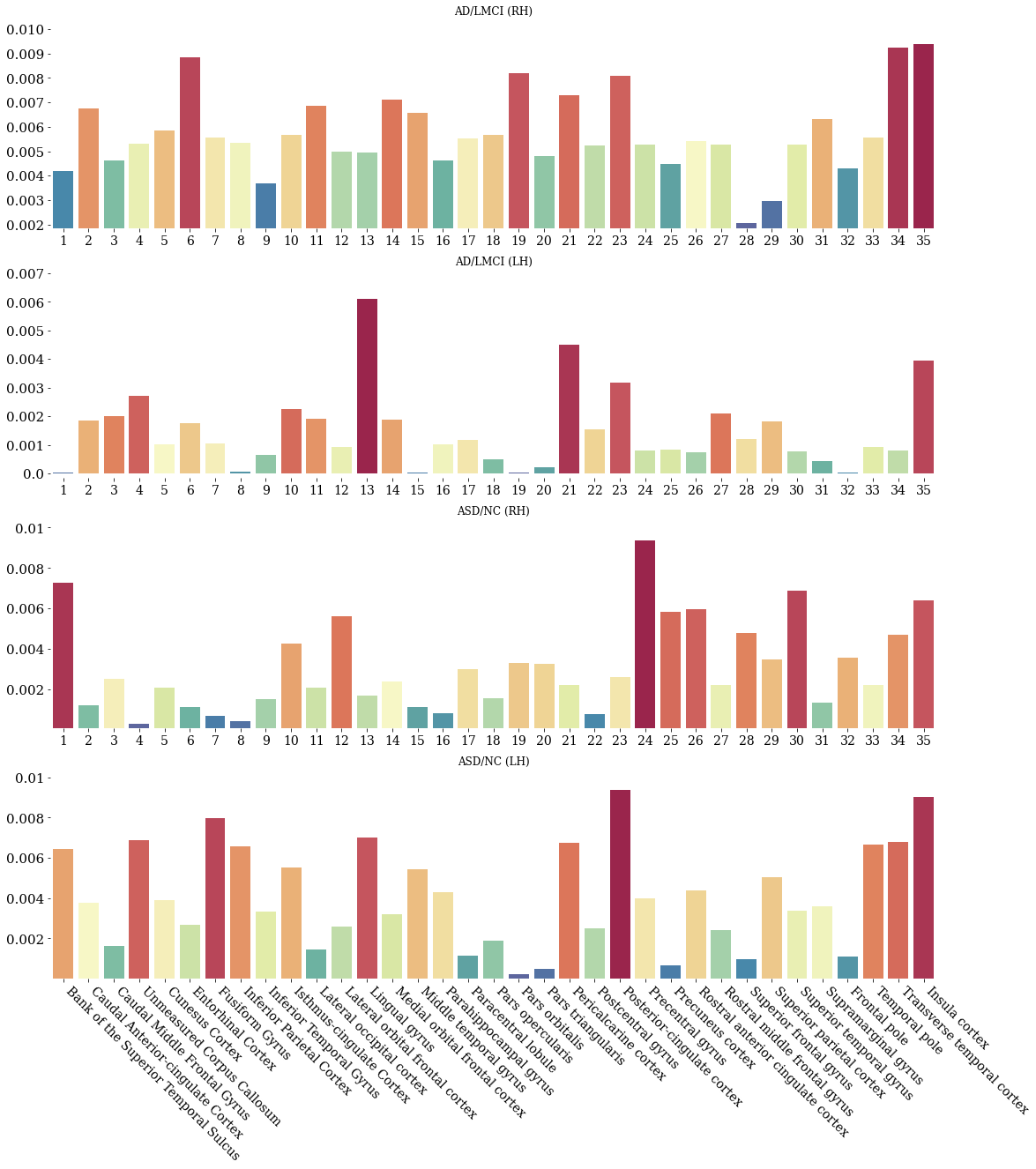}
\caption{\emph{The learned weights for the cortical regions of the brain by the most reproducible model using the four datasets.} AD: Alzheimer's disease. LMCI: late mild cognitive impairment. ASD: autism spectrum disorder. NC: normal control. RH: right hemisphere. LH: left hemisphere. CV: cross-validation. FS: few-shot.}
\label{fig:bars_all}
\end{figure}

For each neurological dataset, we trained GNN models using two different training modes: CV and FS. For CV, we average  3-fold, 5-fold and 10-fold results. For FS, we average the 100 different randomizations that we have conducted to select the few training samples. \textbf{Fig.} \ref{fig:hmaps_adlmcirh} and \ref{fig:hmaps_adlmcilh} illustrate the reproducibility matrices for AD/LMCI RH and LH datasets, respectively. For these datasets, the most reproducible GNNs are DiffPool and SAGPool, respectively. We also note that the most reproducible method is the same across training modes. \textbf{Fig.} \ref{fig:hmaps_asdncrh} and \textbf{Fig.} \ref{fig:hmaps_asdnclh} illustrate the reproducibility matrices for ASD/NC datasets. Based on the overall matrices, g-U-Nets and GAT are the most reproducible models on LH and RH, respectively. For all datasets, the results show that the most reproducible model selection is generalized over different training modes. This emphasizes the robustness of our framework in reproducibility assessment across different data distribution perturbation strategies. In addition, the model having the highest node strength might not be the same across reproducibility scores (correlation-based and average-based). This reflects that the GNN selection highly depends on the reproducibility score. However, the summation of the resulting matrices gave consistent and robust conclusions regarding the most reproducible model selection. Once the most reproducible GNN model is selected, we extract its learned weights as in \textbf{Fig.} \ref{fig:bars_all}. The most discriminative biomarkers will be further detailed in the discussion section.

\section{Discussion}

In this study, we proposed RG-Select to evaluate the reproducibility of GNN models. RG-Select quantifies the reproducibility of a model based on the consensus of its most discriminative biomarkers across other models in a given pool of GNNs. Based on the node strength concept from graph theory, our framework quantifies the reproducibility score of GNNs. In contrast with other methods, RG-Select is applicable to datasets of multigraphs which indicates the challenging level of heterogeneity that can be handled by our framework. This also reflects the generalizability of the settings that we considered in our framework to determine the most reproducible model. Our framework succeeded in producing replicable results across different training modes in all the datasets. In more detail, for each dataset, the most reproducible method was the same across different training settings. Consequently, it identifies the most reliable biomarkers as the most congruent features across models. 

\subsection{Most reproducible biomarkers}
\textbf{Fig.} \ref{fig:bars_all} displays the absolute value of weights respective to biomarkers learned by the most reproducible model for datasets AD/LMCI RH, AD/LMCI LH, ASD/NC RH and ASD/NC LH, respectively. Our framework identified DiffPool as the most reproducible method for AD/LMCI RH dataset as illustrated in \textbf{Fig.}~\ref{fig:bars_all}. The most discriminative ROIs are the lingual gyrus and pericalcarine cortex. Supporting our findings, \citep{shi2020retinal} found that lingual gyrus plays an important role in the neuropathophysiology of depression in AD. Furthermore, \citep{yang2019study} found that pericalcarine cortex was among the brain regions which have a significant reduction in cortical thickness with AD patients. For AD/LMCI LH dataset, SAGPool was selected as the most reproducible model as illustrated in \textbf{Fig} \ref{fig:hmaps_adlmcilh}. The most important biomarkers in this dataset as illustrated in \ref{fig:bars_all} are insula cortex and transverse temporal cortex. \citep{lou2021quantitative} found that the insula cortex has a significant variation in T1 values with AD patients. In addition, \citep{barnes1991angiofensin} showed that the density of the selective angiotensin converting enzyme in temporal cortex is significantly higher with AD patients. For ASD/NC RH dataset, GAT was the most reproducible model. Following the model training, the most two discriminative regions were precentral gyrus and bank of the superior temporal sulcus. \citep{nebel2014precentral} found that the precentral gyrus is highly related to the severity of ASD traits in brain connectivity network. Moreover, \citep{zilbovicius2006autism} stated different experiments showing abnormal or absent superior temporal sulcus activation in patients with ASD during tasks involving social cognition. For the ASD/NC LH dataset, g-U-Nets was the most reproducible GNN model. The experiments have shown that posterior-cingulate cortex and insula cortex are the most discriminative biomarkers. \citep{gogolla2017insular} confirmed that irregularities in insula cortex connectivities are linked to autistic symptom severity. In addition, different studies showed that Abnormalities in posterior-cingulate cortex responses during interpersonal interaction highly correlate with the severity of patients’ autistic symptoms \citep{leech2014role, chiu2008self}.

\subsection{Reproducibility evaluation scores}
\textbf{Tables} \ref{tab:adlmcilh}, \ref{tab:adlmcirh}, \ref{tab:asdnclh} and \ref{tab:asdncrh} contain all the reproducibility scores for all the models on each dataset. We denote the views average and rank correlation based reproducibility scores as v.a and r.c, respectively. Here, we detail the other reproducibility scores not mentioned in the methods section. The first score is the strength correlation (s.c). We extract the node weights of the GNN-specific reproducibility matrix for each model. Then, we compute their average over all the thresholds. The dimension of the resulting vector is equal to $n_v$. Finally, we calculate the correlation score of each pair of GNN resulting vectors. The second score is the accumulated weights correlation (a.w.c). instead of averaging over the thresholds, here, we accumulate them within one vector. The dimension of the resulting vector is $n_v \times n_k$. Next, for each pair of GNNs, we compute the correlation score between their respective resulting vectors. The third score is the accumulated weighted intersection (a.w.i): We calculate the accumulated vectors for each GNN. Next, instead of the conventional intersection, we implement a weighted intersection of the resulting vectors. The weighted intersection takes into consideration of two vectors: accumulated strengths and accumulated rankings. The accumulated strengths vector is the same as the previous method. The accumulated ranking vector contains the rankings of the views (e.g., node strengths of the GNN-specific reproducibility graph). This vector mainly represents the similarities between the strengths and weighted by the similarities in the rankings. In other terms, it gives high scores to elements having close rankings and close strengths. It would also penalize the pair of vectors elements if the elements have close strengths but different ranks. The fourth score is the accumulated rank intersection (a.r.i). It accumulates the vectors of biomarkers at different thresholding based on the GNN-specific reproducibility matrix. Next, we rank the views for each GNN. Eventually, we calculate the correlation between each pair of GNN resulting ranking vectors. The fifth score is the KL divergence (KL). We calculate the ranking vectors of each GNNs as the previous methods. Next, we calculate the KL divergence of the resulting vectors. Unlike previous methods, this score reflects the dissimilarity between both distributions. Therefore, we are interested in identifying the model having the smallest score. Finally, we have the $L2$ distance score (L2) which is constructed by calculating the $L_2$ distance between pairs of the resulting vectors. This score reflects the dissimilarity between GNNs. Hence, based on this score the model having the smallest value is identified as the most reproducible GNN. 

If we consider each reproducibility score independently, the model selection conclusions will be divergent. This also emphasizes the importance and robustness of the two reproducibility scores that we have chosen in our framework. For instance, if we focus on \textbf{Table} \ref{tab:adlmcilh}, each score identifies a different GNN as the most reproducible method. However, the majority of scores indicate that DiffPool is the most reproducible method which confirms our findings for AD/LMCI left hemisphere dataset in \textbf{Fig.} \ref{fig:hmaps_adlmcilh}. For the right hemisphere of the same dataset, the majority of scores in \textbf{Table} \ref{tab:adlmcirh} indicates that SAGPool is the most reproducible model which is the same method selected by our framework for this dataset as detailed in \textbf{Fig.} \ref{fig:hmaps_adlmcirh}. In \textbf{Fig.} \ref{fig:hmaps_asdnclh}, our framework selected g-U-Nets as the most reproducible model for ASD/NC left hemisphere dataset. The same result is confirmed by the majority of the reproducibility scores illustrated in \textbf{Table} \ref{tab:asdnclh}. Finally, the majority of reproducibility scores in \textbf{Table} \ref{tab:asdncrh} reflects that GAT is the most reproducible model which is correlated with our findings in \textbf{Fig.} \ref{fig:hmaps_asdncrh}. 

\begin{table}[H] 
\centering
\tiny
\begin{tabular}{c|c|c|c|c|c|c|c|c|c} 
\hline
Method                    & Training & v.a   & r.c   & a.w.i & a.w.c    & s.c    & a.r.i & KL    & L2     \\ 
\hline
\multirow{2}{*}{DiffPool \citep{ying2018hierarchical}} & CV       & 0.354          & \textbf{0.283} & 0.314          & 0.264          & \textbf{0.238} & 0.255          & \textbf{1.575} & \textbf{2.624}  \\ 
                          & FS       & 0.366          & \textbf{0.25}  & \textbf{0.314} & 0.103          & \textbf{0.32}  & \textbf{0.266} & \textbf{1.782} & 2.809           \\ 
\hline
\multirow{2}{*}{GAT \citep{velivckovic2017graph}}      & CV       & 0.327          & 0.1            & 0.284          & 0.397          & 0.042          & 0.24           & 3.259          & 3.819           \\ 
                          & FS       & 0.334          & 0.1            & 0.247          & \textbf{0.422} & -0.013         & 0.156          & 4.109          & 4.119           \\ 
\hline
\multirow{2}{*}{GCN \citep{kipf2016semi}}      & CV       & 0.353          & 0.083          & \textbf{0.329} & 0.302          & 0.04           & 0.219          & 1.695          & \textbf{2.647}  \\ 
                          & FS       & 0.303          & -0.05          & 0.302          & 0.205          & 0.027          & 0.188          & 2.345          & 3.017           \\ 
\hline
\multirow{2}{*}{g-U-Nets \citep{gao2019graph}} & CV       & \textbf{0.373} & -0.117         & 0.272          & 0.262          & -0.008         & \textbf{0.286} & 2.728          & 3.45            \\ 
                          & FS       & \textbf{0.375} & -0.7           & 0.253          & 0.223          & -0.496         & 0.25           & 3.244          & 3.746           \\ 
\hline
\multirow{2}{*}{SAGPool \citep{lee2019self}}  & CV       & 0.345          & -0.15          & 0.274          & \textbf{0.482} & 0.087          & 0.26           & 2.749          & 3.524           \\ 
                          & FS       & 0.342          & 0.2            & 0.299          & 0.396          & 0.298          & 0.234          & 2.926          & 3.58            \\
\hline
\end{tabular}
\caption{\label{tab:adlmcilh} \emph{Reproducibility scores for AD/LMCI left hemisphere dataset using different GNN models.} AD: Alzheimer's disease. LMCI: late mild congnitive impairment. CV: cross-validation. FS: few-shot. v.a: views average. r.c: rank correlation. a.w.i: accumulated weighted intersection. a.w.c: accumulated weights correlation. s.c: strength correlation. a.r.i: accumulated rank intersection. KL: KL divergence. L2: $L_2$ distance between vectors of scores.}
\end{table}

\begin{table}[H]
\centering
\tiny
\begin{tabular}{l|l|l|l|l|l|l|l|l|l} 
\hline
Method                    & Training & v.a   & r.c   & a.w.i & a.w.c    & s.c    & a.r.i & KL    & L2     \\ 
\hline
\multirow{2}{*}{DiffPool \citep{ying2018hierarchical}} & CV       & 0.343 & -0.05 & 0.345 & \textbf{0.109}  & -0.013 & 0.281 & \textbf{1.499} & \textbf{2.484}  \\ 
                          & FS       & 0.363 & -0.3  & 0.297 & 0.145  & -0.257 & 0.219 & \textbf{1.023} & \textbf{2.39}   \\ 
\hline
\multirow{2}{*}{GAT \citep{velivckovic2017graph}}      & CV       & 0.353 & 0.033 & 0.265 & 0.047  & -0.049 & 0.172 & 2.703 & 3.672  \\ 
                          & FS       & 0.353 & -0.2  & 0.246 & 0.176  & -0.4   & 0.234 & 1.961 & 3.238  \\ 
\hline
\multirow{2}{*}{GCN \citep{kipf2016semi}}      & CV       & 0.342 & 0.033 & \textbf{0.364} & -0.215 & -0.004 & \textbf{0.292} & 1.986 & 2.769  \\ 
                          & FS       & 0.319 & 0.1   & \textbf{0.347} & -0.59  & 0.113  & \textbf{0.266} & 1.777 & 2.848  \\ 
\hline
\multirow{2}{*}{g-U-Nets \citep{gao2019graph}} & CV       & 0.356 & 0     & 0.303 & 0.027  & -0.113 & 0.224 & 2.017 & 2.822  \\ 
                          & FS       & \textbf{0.373} & -0.15 & 0.261 & 0.161  & -0.181 & 0.234 & 1.748 & 2.804  \\ 
\hline
\multirow{2}{*}{SAGPool \citep{lee2019self}}  & CV       & \textbf{0.377} & \textbf{0.05}  & 0.298 & 0.079  & \textbf{-0.001} & 0.271 & 3.129 & 3.944  \\ 
                          & FS       & 0.363 & \textbf{0.15}  & 0.28  & \textbf{0.246}  & \textbf{0.095}  & 0.266 & 2.581 & 3.597  \\
\hline
\end{tabular}
\caption{\label{tab:adlmcirh} \emph{Reproducibility scores for AD/LMCI right hemisphere dataset using different GNN models.} AD: Alzheimer's disease. LMCI: late mild cognitive impairment. CV: cross-validation. FS: few-shot. v.a: views average. r.c: rank correlation. a.w.i: accumulated weighted intersection. a.w.c: accumulated weights correlation. s.c: strength correlation. a.r.i: accumulated rank intersection. KL: KL divergence. L2: $L_2$ distance between vectors of scores.}
\end{table}

\begin{table}[H]
\centering
\tiny
\begin{tabular}{l|l|l|l|l|l|l|l|l|l} 
\hline
method & training & v.a & r.c & a.w.i & a.w.c & s.c            & a.r.i          & KL             & L2              \\ 
\hline
\multirow{2}{*}{DiffPool \citep{ying2018hierarchical}} & CV       & \textbf{0.375} & 0.086          & \textbf{0.346} & 0.415          & 0.01           & 0.194          & 6.901          & 5.779           \\ 
                          & FS       & 0.344          & 0.029          & \textbf{0.383} & 0.52           & 0.002          & \textbf{0.24}  & 4.511          & \textbf{4.622}  \\ 
\hline
\multirow{2}{*}{GAT \citep{velivckovic2017graph}}      & CV       & 0.369          & 0.043          & 0.277          & \textbf{0.627} & 0.095          & 0.208          & 12.879         & 8.605           \\ 
                          & FS       & \textbf{0.38}  & 0.143          & 0.264          & \textbf{0.63}  & 0.225          & 0.188          & 12.574         & 8.853           \\ 
\hline
\multirow{2}{*}{GCN \citep{kipf2016semi}}      & CV       & \textbf{0.375} & -0.029         & 0.334          & 0.361          & 0.028          & 0.17           & \textbf{5.618} & \textbf{5.381}  \\ 
                          & FS       & 0.37           & 0.1            & 0.375          & 0.27           & 0.393          & 0.219          & 4.609          & 4.693           \\ 
\hline
\multirow{2}{*}{g-U-Nets \citep{gao2019graph}} & CV       & 0.372          & \textbf{0.214} & 0.315          & 0.416          & \textbf{0.216} & \textbf{0.222} & 8.232          & 6.456           \\ 
                          & FS       & 0.346          & \textbf{0.414} & 0.339          & 0.476          & \textbf{0.436} & \textbf{0.24}  & 5.632          & 5.306           \\ 
\hline
\multirow{2}{*}{SAGPool \citep{lee2019self}}  & CV       & 0.365          & -0.048         & 0.31           & 0.559          & -0.13          & 0.17           & 8.239          & 6.929           \\ 
                          & FS       & 0.353          & 0.086          & 0.361          & 0.572          & 0.043          & 0.177          & \textbf{4.136} & 5.384           \\
\hline
\end{tabular}
\caption{\label{tab:asdnclh} \emph{Reproducibility scores for ASD/NC left hemisphere dataset using different GNN models.} ASD: autism spectrum disorder. NC: normal control. CV: cross-validation. FS: few-shot. v.a: views average. r.c: rank correlation. a.w.i: accumulated weighted intersection. a.w.c: accumulated weights correlation. s.c: strength correlation. a.r.i: accumulated rank intersection. KL: KL divergence. L2: $L_2$ distance between vectors of scores.}
\end{table}

\begin{table}[H]
\centering
\tiny
\begin{tabular}{l|l|l|l|l|l|l|l|l|l} 
\hline
Method                    & Training & v.a   & r.c    & a.w.i & a.w.c   & s.c    & a.r.i & KL     & L2     \\ 
\hline
\multirow{2}{*}{DiffPool \citep{ying2018hierarchical}} & CV       & 0.361 & 0      & \textbf{0.341} & 0.37  & 0.151  & 0.198 & 6.93   & \textbf{5.788}  \\ 
                          & FS       & 0.35  & 0.186  & \textbf{0.377} & 0.52  & 0.234  & 0.177 & 4.441  & \textbf{4.52}   \\ 
\hline
\multirow{2}{*}{GAT \citep{velivckovic2017graph}}      & CV       & 0.355 & \textbf{0.138}  & 0.281 & \textbf{0.624} & \textbf{0.2}    & \textbf{0.208} & 15.025 & 9.07   \\ 
                          & FS       & 0.378 & \textbf{0.529}  & 0.244 & \textbf{0.627} & \textbf{0.464}  & \textbf{0.229} & 15.764 & 9.645  \\ 
\hline
\multirow{2}{*}{GCN \citep{kipf2016semi}}      & CV       & 0.372 & -0.086 & 0.337 & 0.356 & -0.078 & 0.153 & \textbf{6.513}  & 5.777  \\ 
                          & FS       & \textbf{0.381} & 0.143  & 0.372 & 0.099 & 0.22   & 0.156 & 6.583  & 5.559  \\ 
\hline
\multirow{2}{*}{g-U-Nets \citep{gao2019graph}} & CV       & \textbf{0.373} & 0.005  & 0.294 & 0.342 & -0.036 & 0.132 & 8.931  & 6.694  \\ 
                          & FS       & 0.358 & 0.386  & 0.333 & 0.401 & 0.147  & 0.146 & 5.779  & 5.142  \\ 
\hline
\multirow{2}{*}{SAGPool \citep{lee2019self}}  & CV       & 0.362 & 0.057  & 0.303 & 0.562 & 0.045  & 0.177 & 7.726  & 6.666  \\ 
                          & FS       & 0.37  & 0.129  & 0.321 & 0.516 & 0.091  & 0.188 & \textbf{3.957}  & 5.034  \\
\hline
\end{tabular}
\caption{\label{tab:asdncrh} \emph{Reproducibility scores for ASD/NC right hemisphere dataset using different GNN models.} ASD: autism spectrum disorder. NC: normal control. CV: cross-validation. FS: few-shot. v.a: views average. r.c: rank correlation. a.w.i: accumulated weighted intersection. a.w.c: accumulated weights correlation. s.c: strength correlation. a.r.i: accumulated rank intersection. KL: KL divergence. L2: $L_2$ distance between vectors of scores.}
\end{table}

\subsection{Limitations and future directions}
Although our RG-Select successfully identifies the \emph{most reproducible} graph neural architecture in a given pool of GNNs for a target  multigraph classification task, it has a few limitations. First, our model does not identify the most reproducible view-specific biomarkers since we conceptualized the reproducibility paradigm as finding the model that produces the same biomarkers across different data views \emph{and} various perturbation strategies. However, since a graph is regarded as a special instance of a multigraph where the number of node-to-node edges is equal to 1, one can directly use the GNN-specific reproducibility matrix to first identify the most reproducible GNN then the most reproducible biomarkers. Second, this study only focuses on GNN reproducibility while somewhat overlooking its learning performance in terms of classification accuracy. In our future work, we will investigate the trade-off between GNN reproducibility and performance in different classification tasks. Finally, although we used 5 models for the classification, we have only trained our GNN-based models in a fully supervised manner. As an extension of our RG-Select, we intend to encompass other families of classification methods including semi-supervised and weakly deep learning models.


\section{Conclusion}
While the majority of classification models have focused on boosting the accuracy of a given model, in this study, we address the problem of feature reproducibility. To the best of our knowledge, this is the first work investigating the reproducibility of GNNs in biomarkers using multigraph brain connectivity datasets. Our RG-Select demonstrated robust and consistent results against different training strategies: cross-validation and a few-shot learning. Moreover, we evaluated our framework on both small-scale and large-scale datasets. This work presents a big stride in precision medicine since it incorporates the reproducibility of neurological biomarkers against different perturbations of multi-view clinical datasets. We believe that reproducibility frameworks can make major contributions towards unifying clinical interpretation by enhancing the robustness of the set of biomarkers responsible for brain connectivity alterations in neurologically disordered populations. One major drawback of our framework is the computational time consumed to run all the experiments. To circumvent this issue, we aim, in the foreseeable future, to predict the influence of different perturbations on the overall reproducibility of a given model instead of running it on all datasets. 

\section*{Data Availability}
The data that support the findings of this study are publicly available from ADNI data (\url{http://adni.loni.usc.edu/}). For reproducibility and comparability, the authors will make available upon request all morphological networks generated based on the four cortical attributes (maximum principal curvature, cortical thickness, sulcal depth, and average curvature) for the 77 subjects (41 AD and 36 LMCI) following the approval by ADNI Consortium. Our large-scale dataset is also available from the public ABIDE initiative (\url{http://fcon\_1000.projects.nitrc.org/indi/abide/}). Following the approval by the ABIDE initiative, all morphological networks generated from the six cortical attributes (cortical surface area and minimum principle area in addition to 4 aforementioned measures) for the 300 subjects (150 NC and 150 ASD) are also accessible from the authors upon request.

\section*{Code Availability}

An open-source Python implementation of RG-Select is available on GitHub at \url{https://github.com /basiralab/RG-Select}. The release includes a tutorial, notes regarding Python packages, which need to be installed. Information regarding input format can be also found in the same repository. Input files contain the learned weights by different GNNs. However, the framework works with any data respecting the same shape of the weights vectors extracted from the GNNs.


\section*{Competing Interests}
The authors declare no competing interests.

\section{Acknowledgements}

This work was funded by generous grants from the European H2020 Marie Sklodowska-Curie action (grant no. 101003403, \url{http://basira-lab.com/normnets/}) to I.R. and the Scientific and Technological Research Council of Turkey to I.R. under the TUBITAK 2232 Fellowship for Outstanding Researchers (no. 118C288, \url{http://basira-lab.com/reprime/}). However, all scientific contributions made in this project are owned and approved solely by the authors. M.A.G. is supported by TUBITAK.

Data collection and sharing for this project was funded by the Alzheimer's Disease Neuroimaging Initiative (ADNI) (National Institutes of Health Grant U01 AG024904) and DOD ADNI (Department of Defense award number W81XWH-12-2-0012). ADNI is funded by the National Institute on Aging, the National Institute of Biomedical Imaging and Bioengineering, and through generous contributions from the following: AbbVie, Alzheimer's Association; Alzheimer's Drug Discovery Foundation; Araclon Biotech; BioClinica, Inc.; Biogen; Bristol-Myers Squibb Company; CereSpir, Inc.; Cogstate; Eisai Inc.; Elan Pharmaceuticals, Inc.; Eli Lilly and Company; EuroImmun; F. Hoffmann-La Roche Ltd and its affiliated company Genentech, Inc.; Fujirebio; GE Healthcare; IXICO Ltd.; Janssen Alzheimer Immunotherapy Research \& Development, LLC.; Johnson \& Johnson Pharmaceutical Research \& Development LLC.; Lumosity; Lundbeck; Merck \& Co., Inc.; Meso Scale Diagnostics, LLC.; NeuroRx Research; Neurotrack Technologies; Novartis Pharmaceuticals Corporation; Pfizer Inc.; Piramal Imaging; Servier; Takeda Pharmaceutical Company; and Transition Therapeutics. The Canadian Institutes of Health Research is providing funds to support ADNI clinical sites in Canada. Private sector contributions are facilitated by the Foundation for the National Institutes of Health (www.fnih.org). The grantee organization is the Northern California Institute for Research and Education, and the study is coordinated by the Alzheimer's Therapeutic Research Institute at the University of Southern California. ADNI data are disseminated by the Laboratory for Neuro-Imaging at the University of Southern California.

\newpage
\bibliography{Biblio3}
\bibliographystyle{model2-names}
\end{document}